%% file: main.tex
\documentclass[9pt,technote]{IEEEtran}  
\usepackage{amsmath,amsfonts}
\usepackage{algorithm, algpseudocode}
\usepackage{array}
\usepackage[caption=false,font=normalsize,labelfont=sf,textfont=sf]{subfig}
\usepackage{textcomp}
\usepackage{stfloats}
\usepackage{url}
\usepackage{verbatim}
\usepackage{graphicx}
\usepackage{cite}
\usepackage{xcolor}
\usepackage{booktabs}
\begin{document}

\title{Applying Graph Explanation to Operator Fusion}

\author{Keith G. Mills\thanks{K. G. Mills, M. Fetrat Qharabagh, W. Qiu and D. Niu are with the Department of Electrical and Computer Engineering, University of Alberta at Edmonton, Alberta, Canada, T6G 1H9}, Muhammad Fetrat Qharabagh, Weichen Qiu, \\Fred X. Han\thanks{F. X. Han, M. Salameh, and W. Lu are with Huawei Technologies, Edmonton, Canada, T6G 2C8}, Mohammad Salameh, Wei Lu, Shangling Jui\thanks{S. Jui is with Huawei Kirin Solution, Shanghai, China, 200120}, Di Niu}

\maketitle

\input{src/abstract}

\input{src/introduction}
\input{src/background}
\input{src/method}
\input{src/results}
\input{src/conclusion}

\bibliography{ref}
\bibliographystyle{IEEEtran}

\vfill

\end{document}

%% file: src/abstract.tex
\begin{abstract}

Layer fusion techniques are critical to improving the inference efficiency of deep neural networks (DNN) for deployment. Fusion aims to lower inference costs by reducing data transactions between an accelerator's on-chip buffer and DRAM. This is accomplished by grouped execution of multiple operations like convolution and activations together into single execution units - fusion groups. However, on-chip buffer capacity limits fusion group size and optimizing fusion on whole DNNs requires partitioning into multiple fusion groups. Finding the optimal groups is a complex problem where the presence of invalid solutions hampers traditional search algorithms and demands robust approaches. 

In this paper we incorporate Explainable AI, specifically Graph Explanation Techniques (GET), into layer fusion. Given an invalid fusion group, we identify the operations most responsible for group invalidity, then use this knowledge to recursively split the original fusion group via a greedy tree-based algorithm to minimize DRAM access. We pair our scheme with common algorithms and optimize DNNs on two types of layer fusion: Line-Buffer Depth First (LBDF) and Branch Requirement Reduction (BRR). Experiments demonstrate the efficacy of our scheme on several popular and classical convolutional neural networks like ResNets and MobileNets. Our scheme achieves over 20\% DRAM Access reduction on EfficientNet-B3.

\end{abstract}

%% file: src/introduction.tex
\section{Introduction}
\label{sec:intro}

Deep Neural Networks (DNN) have become an indispensable tool when applying machine learning techniques to solve real-world problems such as computer vision tasks. Performance gains are usually observed as the network becomes larger and deeper. The upsurge in hardware computational power and throughput, such as GPUs and TPUs, is sustaining this performance, thus enabling the training of larger DNNs in shorter periods. Concurrent with these advances, there is a need to develop methods that will allow faster DNN inference for deployment on downstream hardware accelerators.

Layer Fusion (LF; also known as Operator Fusion)~\cite{sze2017efficient, cai2021optimus, zheng2020efficient, shi2021hardware, boehn2018Operator} is a form of inference acceleration that aims to reduce the number of data transactions between the on-chip buffer of a neural accelerator and a corresponding off-chip DRAM, as each transaction is costly in terms of power and inference latency. To facilitate this, the execution of multiple DNN operation layers, e.g., Convolutions and ReLU, are \textit{fused} together to reduce the amount of intermediate data that must be written back to the DRAM.

Layer Fusion optimization first casts a DNN as a Directed Acyclic Graph (DAG) where each node corresponds to a single operation layer, e.g., convolution instance, while the edges are defined by the DNN forward pass. This DAG is then partitioned into \textit{fusion groups}. Each fusion group is a subgraph of operations that execute according to an efficient LF scheduling pattern~\cite{sze2017efficient, boehn2018Operator} such as Line Buffer Depth-First (LBDF) execution~\cite{shi2021hardware} and Buffer Requirement Reduction (BRR)~\cite{cai2021optimus}. A \textit{partition plan} is a supergraph consisting of all fusion groups that determines the overall inference execution by mapping the flow of data tensors between fusion groups. Since the only intermediate results that need to be transferred to and from the DRAM are the inputs and outputs of each fusion group, the total number of DRAM transactions is reduced, which in turn lowers inference energy cost and latency delay.

The data and weight requirements vary with each LF type. For instance, LBDF executes a series of stacked convolutions at once. As such, the weights of each convolution operation in an LBDF fusion group must be stored within the on-chip buffer for the entire execution of the fusion group. By contrast, BRR allows parameter-induced DRAM access by partitioning the weights into sub-groups for sequential execution. Regardless of these specifics, each fusion group has an associated minimum memory size required to execute. Moreover, the fixed size of a given accelerator's on-chip buffer imposes a hard constraint on the feasibility of LF methods. If the memory requirements of a fusion group exceed buffer capacity, that fusion group cannot execute.

Additionally, since modern DNNs can contain hundreds of layers~\cite{sandler2018mobilenetv2, howard2019searching, tan2019efficientnet}, it is generally infeasible to fuse all of them together. Instead, finding partition plans with low DRAM access is a combinatorial optimization over an ample search space. Although search approaches like Evolutionary Algorithms (EA)~\cite{liu2021EvoSurvey}, Local Search (LS)~\cite{stutzle1999local}, and even Random Search (RS)~\cite{li2020random} are effective optimization tools for large search spaces, the constraint imposed by on-chip buffer size hampers effectiveness. Suppose a search algorithm proposes a partition plan that consists of an invalid fusion group. That group must be split, imposing additional DRAM costs, or the entire partition plan is invalid. Moreover, the search generally executes using a fixed budget quantified by the number of partition plans an algorithm can propose. Having an inefficient search algorithm that generates many invalid partition plans or suboptimal fusion groups is not desirable.

We address these concerns and step toward more robust search algorithms for LF. We propose to use Graph Explanation Techniques (GET) to resolve invalid fusion groups intelligently. Specifically, given a graph object and corresponding Graph Neural Network (GNN), a GET will find the subgraph of nodes and edges that the GNN relies upon most when making a prediction on the original graph. In this paper we apply GET to LF in order to find partition plans with low DRAM cost. Our detailed contributions are as follows:

First, we cast the problem presented by invalid fusion groups as a recursive optimization task. While an invalid fusion group can be randomly partitioned into smaller, valid groups, the DRAM access of these new fusion groups may not be optimal. Therefore, we develop a tree-based partitioning scheme for rectifying invalid fusion groups. It incorporates recursion and greedy logic to find solutions with low DRAM access cost.

Second, we cast the process of determining a fusion group's validity as a binary classification problem. We consider several prominent GETs, such as GNNExplainer~\cite{ying2019gnnexplainer}, PGExplainer~\cite{luo2020parameterized} and RG-Explainer~\cite{shan2021reinforcement} to discover the subgraph nodes and edges responsible for rendering a given fusion group invalid. We pass this information on to our partition scheme to intelligently and surgically split the fusion group.

We incorporate our scheme with several search algorithms, such as Local Search, Random Search, and NSGA-II~\cite{yusoff2011overview}, to demonstrate how it can find network partition plans with lower DRAM cost. Specifically, we consider two types of LF, namely BRR and LBDF across varying on-chip buffer sizes. Furthermore, to illustrate the relevance of our scheme, we experiment across a spectrum of modern and classical DNN designs ranging from EfficientNets~\cite{tan2019efficientnet}, MobileNetV2 (MBv2)~\cite{sandler2018mobilenetv2}, ResNets~\cite{he2016deep}, SqueezeNet~\cite{iandola2016squeezenet}. To demonstrate the applicability of our scheme across computer vision tasks, we also consider a Semantic Segmentation network, DeepLabV3+MobileNetV3~\cite{deeplabv3plus2018, howard2019searching}. Experimental results demonstrate that our method helps find better partition plans across a range of LF execution schemes, on-chip buffer sizes, and search budgets.

The rest of this paper is organized as follows: We provide a high-level overview of LF and GETs in Section~\ref{sec:background} before elaborating on our proposed scheme in Section~\ref{sec:method}. We provide a detailed experimental setup and results in Section~\ref{sec:results} before concluding in Section~\ref{sec:conclusion}.

%% file: src/background.tex
\section{Background and Related Work}
\label{sec:background}

The field of Layer Fusion (LF) has become essential and complex~\cite{sze2017efficient} as the design of DNNs grows deeper and more intricate. For example, \cite{waeijen2021convfusion} consider node clustering for irregular network structures with many branches and skip-connections. \cite{shi2021hardware} consider ReLU-based compression and tiling effects to combat the adverse effect that skip-connections can have on LF execution. LF methods can have different levels of specificity. For instance, \cite{boehn2018Operator} speed up general linear algebra operations by exploiting sparsity. Other forms of LF may focus focus on improving the inference costs specific DNN operations, e.g., Convolutional Neural Networks (CNN) or the dense matrix-multiplication-softmax sequences that comprise attention-based models~\cite{dao2022flashattention}.

\begin{figure}
    \centering
    \includegraphics[width=3.4in]{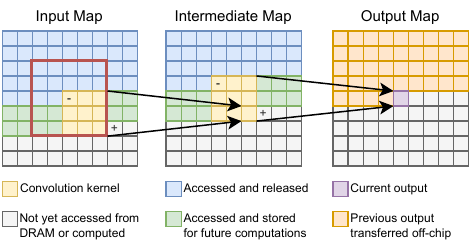}
    \caption{LBDF on a fusion group consisting of two $3 \times 3$ convolution kernels in sequence. Area bounded by the red square denotes the input data required to compute the current output. `-' denotes the next data entries to be released from the on-chip buffer. `+' denotes the next data point to be loaded from DRAM (input map) or computed (intermediate map). Best viewed in color.}
    \label{fig:lbdf_example}
\end{figure}

Specifically, we consider two forms of LF for CNNs: Line-Buffer Depth-First (LBDF)~\cite{shi2021hardware} and Buffer Requirement Reduction (BRR) execution. Figure~\ref{fig:lbdf_example} provides a sample illustration of how LBDF inference is performed. While the weights of all convolution operations must be stored in the on-chip buffer at all times, a sliding window mechanism ensures that only a fraction of the input and intermediate feature map needs to be stored on-chip at any given time. A downside is that if the size of an operator's weight tensor exceeds the size of the on-chip buffer, that operator cannot be used with LBDF and must execute using another inference method.

By contrast, BRR relaxes the requirement that all weight tensors be stored on-chip during inference by allowing parameter-induced memory access. This can be advantageous in cases where a target fusion group contains branching operation paths~\cite{liu2021EvoSurvey, li2020random} that share the same intermediate values as input. A drawback of BRR is that it is designed for classical CNN architectures and is not friendly to newer CNN structures such as the Squeeze-and-Excite (SE)~\cite{hu2018squeeze} module found in MobileNetV3~\cite{howard2019searching} and EfficientNet~\cite{tan2019efficientnet}.

Regardless of scheme, LF is generally cast as a search problem over how to partition a given DNN, with a hard constraint defined by the on-chip buffer size, and other potential hardware restrictions. We provide a generalized overview of the LF optimization problem before introducing Graph Explanation Techniques (GET) in the remainder of this section.

\begin{figure*}
    \centering
    \includegraphics[width=7in]{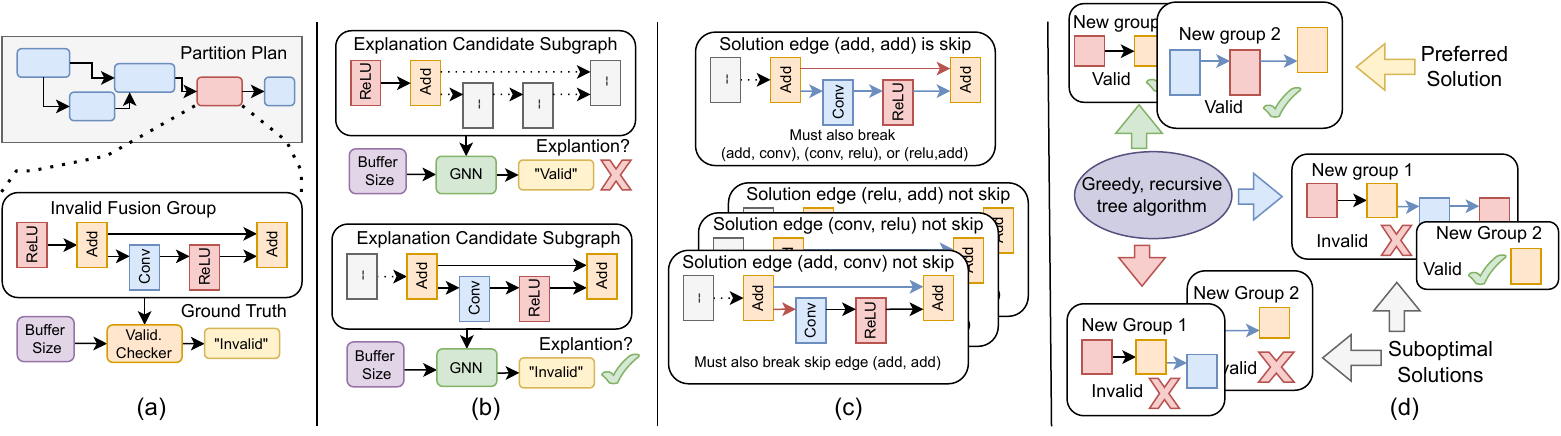}
    \caption{A high-level overview of our scheme. Best viewed in color. (a): A search algorithm generates a partition plan, and an analytical validity checker determines the feasibility of each fusion group in the plan. (b): We use a GNN and GETs to find a subgraph explanation for each invalid fusion group. (c): We consider how to split the fusion group at every solution edge contained within the subgraph explanation. Note how the explanation contains a skip-connection, meaning we must cut at least 2 edges. (d): We use a greedy tree-based algorithm to consider all possible solutions which split the fusion group and sort them based whether the number of new fusion groups. In the optimal case (green arrow), both new fusion groups are valid. If one (blue arrow) or both (red arrow) of the fusion groups are invalid, we use the recursive algorithm to repeat the process from step (a) for each invalid fusion group.}
    \label{fig:fusion_explanation}
\end{figure*}

\subsection{Networks as Graphs}

A common approach to LF is to cast a DNN as a Directed Acyclic Graph (DAG) or Computational Graph (CG). In a CG, each node is a primitive operation, e.g., convolution, concatenation, activation, etc., while edges describe the DNN forward pass. LF optimization is then cast as a graph partition search problem. Let $G$ be a CG with node set $\mathcal{E}_G$ and edge set $\mathcal{V}_G$. LF optimization partitions $G$ into a \textit{partition plan} $\Phi$, which consists of $N$ disjoint \textit{fusion groups} $\mathcal{E}_\Phi = \{\phi_0, \phi_1,..,\phi_{N-1}\}$ and a subset of the original edges, $\mathcal{V}_\Phi \subset \mathcal{V}_G$. 

Fusion groups are disjoint subgraphs of $G$, while $\mathcal{V}_\Phi$ represents the connections \textit{between} fusions groups. Like $G$, $\Phi$ is a DAG that has no cyclic dependencies. Moreover, for $\Phi$ to be valid, each fusion group should be weakly connected; i.e., the underlying undirected graph representation is connected. At inference time, each fusion group will execute separately under an efficient LF scheme, while $\mathcal{V}_\Phi$ represent the intermediate data tensors that transact between the on-chip buffer and off-chip DRAM. 

It is possible to have a simple partition plan where every node in $G$ is its own fusion group ($N = |\mathcal{E}_G|$) and $\mathcal{V}_\Phi = \mathcal{V}_G$. However, this scenario is suboptimal as we need to perform a DRAM-buffer transaction for every edge in $\mathcal{V}_\Phi$ which is costly in terms of latency and power, thus motivating the development of an effective search procedure.

\subsection{Subgraph Explanations}
\label{sec:ge}

Graph Explanation is a recent field of eXplainable AI (XAI) that has been gaining popularity. Graph Explanation Techniques (GET) provide a qualitative interpretation of the predictions Graph Neural Networks (GNN)~\cite{morris2019weisfeiler} make. For example, if a GNN is trained to performed binary classification, GETs aim to identify the graph attributes, e.g., nodes, edges, motifs, etc., which influence the decision-making process.

Formally, let $\theta$ denote the parameters of a given GNN and let $G$ be an input graph. Feeding $G$ into the GNN produces $p(G|\theta)$, the class probability distribution we wish to explain. A GET will formulate an explanation as a subgraph $G^* \in G$ that heavily influence the prediction. As even a small graph can have many subgraph permutations with varying numbers of nodes and edges, GETs need a measure to quantify the importance of any given subgraph. For example, GNNExplainer (GNNE)~\cite{ying2019gnnexplainer} use Mutual Information (MI) to measure importance by finding the subgraph $G^*$ which maximizes the MI between itself and the original graph $G$. More formally, the GNNE objective is denoated as 

\begin{equation}
    \centering
    \label{eq:mi}
    \underset{G^*}{\text{max}} MI(p(G|\theta), G^*) = H(p(G|\theta)) - H(p(G^*|\theta)),
\end{equation}
where $H$ denotes the entropy of the probability distribution $p(.|\theta)$. GNNE solves the MI problem using a fractional adjacency matrix to control the number of subgraph candidates considered, but repeats its explanation process from scratch for every new graph $G$.

PGExplainer (PG)~\cite{luo2020parameterized} extend the concept of GNNE by pre-training a set of parameters on top of the original GNN in order to speedup the downstream explanation process. Specifically, for every edge in a graph, PG concatenates the corresponding node embeddings produced by the GNN and pre-trains a simple MLP which associates the concatenated embeddings with the overall prediction $p(G|\theta)$. This allows PG to identify significant edges and quickly construct $G^*$. 

Since GNNE and PG explicitly focus on the independent importance of nodes/edges, the explanations they produce are not guaranteed to be connected graphs. By contrast, RG-Explainer (RG)~\cite{shan2021reinforcement} is a GET that utilizes Reinforcement Learning (RL) to generate connected subgraph explanations. Like PG, RG pre-trains additional parameters (e.g., an MLP) on top of the target GNN, but then adopts a 3-phase approach to generate explanations: starting point selection, iterative graph generation, and stopping criteria.

Given an input graph, RG uses node embeddings and an MLP to select a \textit{seed node} as the starting point for the explanatory subgraph $G^*$. Then RG uses an agent to iteratively take actions by selecting neighboring nodes to add to the subgraph. The explanation $G^*$ is complete when RG selects a special termination action rather than adding another neighboring node. Finally, RG uses the cross-entropy loss between $p(G|\theta)$ and $p(G^*|\theta)$ to calculate a reward. Additionally, RG considers additional loss penalties based on the subgraph's radius and number of nodes. We refer interested readers to \cite{ying2019gnnexplainer, luo2020parameterized, shan2021reinforcement} for further details on these GETs.

%% file: src/method.tex
\section{Methodology}
\label{sec:method}

At its base level, LF optimization is a graph partition problem that can be solved with different kinds of search algorithms with varying complexities. However, the effectiveness of any search algorithm depends on how well it can handle invalid fusion groups. 

Let $\phi_n$ be an arbitrary fusion group from a candidate partition plan $\phi_n \in \Phi$ that has been produced by a search algorithm. Furthermore, let $\beta$ be the on-chip buffer capacity, let $F_\beta$ be a validator function that queries the buffer memory requirements of a given fusion group and let $F_D$ be a function that computes the DRAM access cost to perform inference on a fusion group. The overall search objective is to find the best partition plan that minimizes DRAM access subject to the on-chip buffer size constraint:

\begin{equation}
\begin{aligned}
    \min_\Phi \sum_{\phi_n \in \Phi} F_D(\phi_n), \\
    \textrm{s.t.} \forall \phi_n \in \Phi \mid F_\beta(\phi) < \beta. \\
    \label{eq:objective}
\end{aligned}
\end{equation}

That is, for $\Phi$ to be valid, the memory cost of each individual fusion group cannot exceed the buffer capacity. Search complexity arises when considering how to process invalid fusion groups. One simple method would be to discard the entire partition plan, and generate a new one. However, this option is suboptimal as some of the other fusion groups in the partition plan could have below average DRAM access cost. A second, but also simple method is to randomly split the invalid fusion group into two or more new fusion groups which might be valid. This is also suboptimal as there is no guarantee that the new fusion groups will have desirable DRAM cost.

We aim to solve this problem by providing an explainable technique for splitting invalid fusion groups. Figure~\ref{fig:fusion_explanation} provides a high-level overview of our proposed scheme. In the following subsections we will iterate across Figure~\ref{fig:fusion_explanation} and elaborate on the details of how our method selects invalid fusion groups, uses GETs to determine invalidity, considers how to split fusion groups with skip (e.g., residual) connections and how a greedy, tree-based method allows us to select the optimal way to perform a split.

\subsection{Cost Model Granularity}

Figure~\ref{fig:fusion_explanation}(a) shows how fusion group validity is determined by calculating if its memory requirements exceed the buffer size, $F_\beta(\phi) < \beta$. At its most base level, $F_\beta$ and $F_D$ represent a cost model implementation of the mathematical equations to compute buffer size for a given LF method given the size of weight and data tensors in a fusion group. More advanced cost models could incorporate hardware specifications.

While these functions can be used to identify fusion group buffer and DRAM access costs, they cannot be directly paired with GETs which are designed to operate on GNNs and require access to the latent representation of graph nodes and edges in order to find an explanation without performing a costly exhaustive search of all possible solutions. Additionally, deriving an explanation from $F_\beta$ directly is not generalizable as the mathematical equations are specific to different types of LF. 

\subsection{Fusion Group Explanation}
We cast the problem of determining fusion group validity as a binary classification problem and train a GNN~\cite{morris2019weisfeiler} to mimic the behaviour of the mathematical validity check,

\begin{equation}
    \centering
    Validity = \sigma(p(y|\phi, \beta, \theta)),
    \label{eq:gnn}
\end{equation}
where $y$ is a discrete `yes/no' on whether $\phi$ is invalid, converted from the continuous probability $p$ by argmax $\sigma$. As shown in Figure~\ref{fig:fusion_explanation}(b), we pass invalid fusion groups to a GNN $\theta$ and GET $\Theta$, e.g., GNNE, PG or RG, to first provide an explanation. $\Theta$ produces a set of edges $\epsilon_{(\phi_n, \beta)} = \{(i, j)\} \in \mathcal{E}_\phi$. Formally, 

\begin{equation}
    \centering
    \label{eq:explanation}
    \epsilon_{(\phi_n, \beta)} = \Theta(\phi_n, \beta, \theta).
\end{equation}

Each edge in $\epsilon_{(\phi_n, \beta)}$ represents a pair of nodes $i$ and $j$ whose layer fusion cost contributes to the invalidity of $\phi_n$. Conceptually, our method involves splitting $\phi_n$ along $(i, j)$ into two new, disconnected fusion groups, $\phi_n^i$ and $\phi_n^j$, and each edge represents a potential solution to consider. However, the presence of skip-connections ensures that sometimes we need to remove more than one edge to split a fusion group.

\subsection{Skip-Connections}
\label{sec:skip}

Modern DNN architectures employ residual skip-connections to improve generalization performance and learning~\cite{he2016deep, hu2018squeeze, sandler2018mobilenetv2, howard2019searching, tan2019efficientnet}. In the context of splitting fusion groups, the use of skip-connections means that removing one edge $(i, j)$ may not be enough to separate the original fusion group into two disconnected subgraphs. To address this concern, we start by topologically sorting every operation node in the original DNN and assigning them an ascending numerical label. If there are $|\mathcal{E}|$ total nodes, the first input is node $0$, the last output is node $|\mathcal{E}| - 1$, and $\forall_{(i, j) \in \mathcal{E}}, i < j$.

If the removal of a given edge $(i, j)$ cannot separate the original fusion group $\phi_n$ into disconnected $\phi_n^i$ and $\phi_n^j$, we consider two scenarios. First, $(i, j)$ represents a skip-connection (upper half of Fig.~\ref{fig:fusion_explanation}(c)), and there are additional edges connecting $\phi_n^i$ and $\phi_n^j$. For each additional edge, we consider whether its removal will separate $\phi_n^i$ and $\phi_n^j$ and augment the original solution $(i, j)$. If $\phi_n^i$ and $\phi_n^j$ are still connected, the subgraph contains nested skip-connections, e.g., the Squeeze-and-Excitation~\cite{hu2018squeeze} modules in EfficientNets~\cite{tan2019efficientnet} and MobileNetV3~\cite{howard2019searching} and we need to consider additional edges to remove recursively.

In the second case (lower part of Fig.~\ref{fig:fusion_explanation}(c)), $(i, j)$ is encompassed by at least one skip-connection. We can identify these skip-connections and remove them as well. To accommodate overlap between identified edges, we maintain the minimal possible set of solutions when considering skip-connections, e.g., the explanation in Figure~\ref{fig:fusion_explanation}(c) only produces three solutions.

\subsection{Greedy Tree-based Selection}

\begin{algorithm}[t]
  \caption{Recursive Greedy Tree-Based Splitting}
  \label{alg:pseudo}
  \begin{algorithmic}[1]
    \Function{Split}{$\phi$, $\beta$, $F_\beta$, $F_D$, $\theta$, $\Theta$}
        \State $\epsilon_{(\phi, \beta)} = \Theta(\phi_n, \beta, \theta)$ \Comment{Includes edges from Sec.~\ref{sec:skip}.}
        \State $\mathcal{S}_1$, $\mathcal{S}_2$, $\mathcal{S}_3$ = $\emptyset$, $\emptyset$, $\emptyset$ \Comment{Three categories}
        \For{$(i, j)_\phi \in \epsilon_{(\phi, \beta)}$}
            \State $\phi_i$, $\phi_j$ = \texttt{Partition}($(i, j)_\phi$, $\phi$)
            \If{$F_\beta(\phi_i) < \beta$ and $F_\beta(\phi_j) < \beta$}
                \State $\mathcal{S}_1$ += $(\phi_i$, $\phi_j)$  \Comment{Preferred solution}
            \ElsIf{$\mathcal{S}_1 = \emptyset$}
                \If{$F_\beta(\phi_i) < \beta$ or $F_\beta(\phi_j) < \beta$}
                \State $\phi_{valid}, \phi_{invalid} = \texttt{Sort}(\phi_i, \phi_j)$
                \State $\mathcal{S}_2$ += $(\phi_{valid}, \phi_{invalid})$ \Comment{Intermediate solution}
                \Else
                \State $\mathcal{S}_3$ += ($\phi_i$, $\phi_j$) \Comment{Worst-case scenario}
                \EndIf
            \EndIf
        \EndFor
        \If{$\mathcal{S}_1 != \emptyset$} \Comment{Preference to preferred solutions (Cat. 1)}
            \State \Return $\texttt{min}(\mathcal{S}_1, F_D)$ \Comment{Minimum DRAM Access}
        \ElsIf{$\mathcal{S}_2 != \emptyset$}
            \State $(\phi_{valid}^*, \phi_{invalid}^*) = \texttt{max\_valid\_nodes}(\mathcal{S}_2)$
            \State \Return $(\phi_{valid}^*, \texttt{Split}$($\phi_{invalid}^*$, $\beta$, $F_\beta$, $F_D$, $\theta$, $\Theta$))
        \Else
            \State $\mathcal{S}_3^*\ = \emptyset$        
            \For{$(\phi_i, \phi_j) \in \mathcal{S}_3$}  \Comment{Split all invalid groups}
                \State ($\phi^*_i$) = \texttt{Split}($\phi_i$, $\beta$, $F_\beta$, $F_D$, $\theta$, $\Theta$)
                \State ($\phi^*_j$) = \texttt{Split}($\phi_j$, $\beta$, $F_\beta$, $F_D$, $\theta$, $\Theta$)
                \State $\mathcal{S}_3^*$ += $((\phi^*_i), (\phi^*_j))$
            \EndFor
            \State \Return $\texttt{min}(\mathcal{S}_3^*, F_D)$
        \EndIf
    \EndFunction
  \end{algorithmic}
\end{algorithm}

Minimizing DRAM access requires considering all possible solutions in $\epsilon_{(\phi_n, \beta)}$. As Figure~\ref{fig:fusion_explanation}(d) shows, these solutions can be coarsely grouped depending on the number of valid fusion groups they produce. In order to iterate across these solutions and select the optimal one, we adopt a recursive tree-based approach and incorporate greedy logic to solve fusion group invalidity. As Algorithm~\ref{alg:pseudo} shows, given an invalid fusion group $\phi$, we first compute the set of solution edges $\epsilon_{(\phi_n, \beta)}$ in line 2\footnote{Including solutions that accomodate skip-connections per Sec.~\ref{sec:skip}.} and then group these solutions into three categories: 

\begin{enumerate}
    \item \textbf{Category 1:} $F_\beta(\phi_n^i) < \beta \land F_\beta(\phi_n^j) < \beta$, considered in lines 6-7, denotes a preferred solution and algorithm end-point. If any solution fits this criterion, we choose the one that minimizes $F_D(\phi_n^i) + F_D(\phi_n^j)$, the combined cost of both new fusion groups. Moreover, once we know at least one category 1 solution exists, we adopt greedy logic and do not even consider solutions that fall into the other two categories (line 8) as they will necessarily require splitting $\phi$ into 3 or more fusion groups and therefore will incur higher DRAM access.
    \item \textbf{Category 2:} $F_\beta(\phi_i) > \beta \lor F_\beta(\phi_j) > \beta$ is an intermediate solution as one of the new fusion groups is invalid, while the other is valid (lines 9-12). If multiple solutions exist in this category, we take a greedy approach (lines 19-21) and select the solution where the valid fusion group contains the most nodes (line 20), then recursively split the invalid fusion group (e.g., repeating the procedure from Fig.~\ref{fig:fusion_explanation}(b) onwards). 
    \item \textbf{Category 3:} $F_\beta(\phi_i) > \beta \land F_\beta(\phi_j) > \beta$ is the worst-case scenario as both fusion groups are still invalid. We only consider solutions of this category when no Category 1 or 2 solutions exist. We run another round of recursion (lines 24-27) on each set of invalid fusion groups $(\phi_i, \phi_j)$, then select the solution which minimizes DRAM access.
\end{enumerate}

It should be noted that if $\phi$ only contains 1 node, $\epsilon_{(\phi, \beta)}$ will necessarily be an empty set $\emptyset$ and default to returning nothing (lines 23 and 29). This represents a case where fusion simply cannot be performed on a given node, e.g., LBDF where the size of an operation weight tensor itself exceeds buffer size, however, we aim to identify such problematic nodes prior to LF optimization and exclude them from search. Overall though, Algorithm~\ref{alg:pseudo} aims to balance the objective of minimizing DRAM access while also minimizing queries to the ground-truth buffer size and DRAM Access profilers, $F_\beta$ and $F_D$, respectively. 

%% file: src/results.tex
\section{Experimental Results}
\label{sec:results}

\begin{table*}[t]
    \centering
    \caption{Number of fusion group samples extracted using random search across numerous CNNs. `MB', `RN', and `EB' denote MobileNets, ResNets and EfficientNets of varying sizes, respectively.}
    \label{tab:data}
    \scalebox{0.95}{
    \begin{tabular}{lccccccccc} \toprule
    \textbf{CNN Architecture} & \textbf{VGG16} & \textbf{SqueezeNet} &\textbf{MBv2} & \textbf{MBv3-Small} & \textbf{MBv3-Large} & \textbf{RN-18} & \textbf{RN-101} & \textbf{EB0} & \textbf{EB3} \\ \midrule
    \#Samples & 715 & 795 & 1100 & 203 & 251 & 715 & 3476 & 1364 & 2247 \\ \bottomrule
    \end{tabular}
    }
\end{table*}

\begin{figure*}
    \centering
    \includegraphics[width=7in]{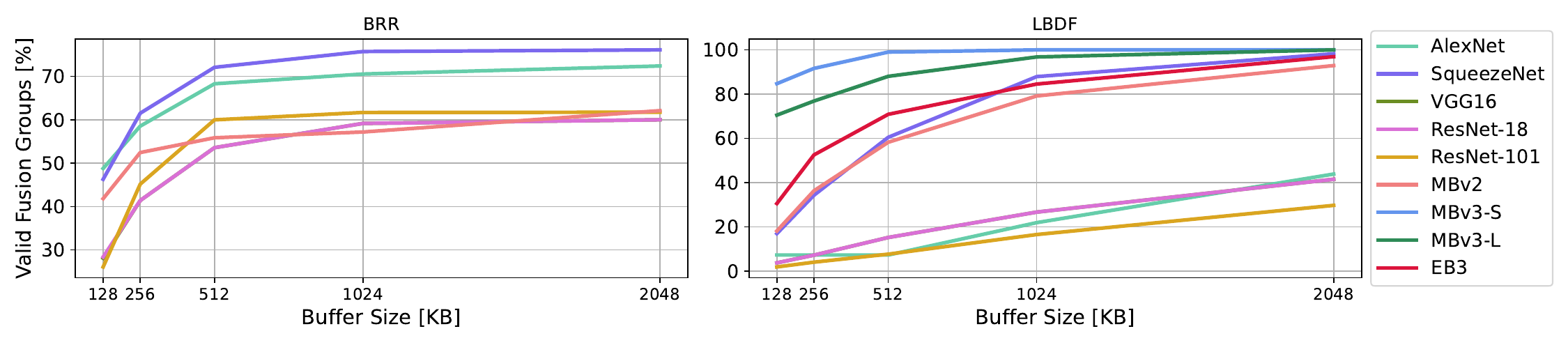}
    \caption{Plotting the percentage of valid fusion groups for each CNN type as we double the on-chip buffer size from a minimum of 128KB to a maximum of 2048KB. Best viewed in color. Note that we do not plot BRR curves for MobileNetV3 and EfficientNet or use them to train our GNNs as that form of LF has compatibility issues with the Squeeze-and-Excite~\cite{hu2018squeeze} module present in those networks.}
    \label{fig:both_plots}
\end{figure*}

In this section we elaborate on our experimental setup, execution and then enumerate our findings. Specifically, we provide details on our scheme for pretraining GNNs and GETs (if applicable) and include data collection statistics. We then iterate our chosen LF schemes and search algorithms, then present results on several modern CNN architectures like EfficientNet~\cite{tan2019efficientnet}, MobileNets~\cite{sandler2018mobilenetv2, howard2019searching} as well as several classical designs like ResNets~\cite{he2016deep}, SqueezeNet~\cite{iandola2016squeezenet} and VGG16~\cite{simonyan2014very}.

\subsection{Graph Neural Networks to Predict Validity}

For each form of LF we consider, we train a GNN binary classifier that can pair with GETs and mimics the behaviour of the LF validity checker through Equation~\ref{eq:gnn}. When testing the effectiveness and generalizability of our proposed scheme, we consider on-chip buffer size as the principle hyperparameter criterion for determining fusion group validity for a given LF scheme. However, we note that our scheme can be paired with more advanced validity checkers that are designed with specific dataflows or hardware constraints.

To train the GNNs, we run a simple LF random search on the ONNX~\cite{bai2019onnx} CG model representations of numerous CNN architectures (e.g., ResNets, MobileNets, SqueezeNets, VGG, etc.) in order to generate a wide breadth of valid and invalid fusion groups of various sizes, topologies and operation combinations, for a total of over 10.8k fusion group subgraphs. We then use the validity checker $F_\beta$ to compute the actual buffer memory cost of each fusion group. Moreover, we multiply the size of our dataset by considering a range of buffer sizes $\beta \in \{128KB, 256KB, 512KB, 1024KB, 2048KB\}$ for each fusion group sample, giving us a total of over 54k samples. When converting ONNX graphs to GNN data, we encode a specific buffer size as a CG node feature alongside operation type, input/output tensor shape and weight tensor size.

Table~\ref{tab:data} provides a breakdown of how many samples we extract from each CNN, while Figure~\ref{fig:both_plots} illustrates the proportion of fusion groups that are valid as we increase buffer size from 128KB to 2048KB for both the LBDF and BRR LF contexts. As expected, the number of valid fusion groups increases monotonically with buffer size, though with ever diminishing returns. Moreover, we note the difference in percentage ranges as validity for BRR is between 25\%-80\% but can vary between 0\%-100\% for LBDF, depending on the CNN.

Our GNNs consist of 4 $k$-GNN~\cite{morris2019weisfeiler} layers with a hidden size of 128, ReLU activation and sum aggregation\footnote{Graph embedding is the summation of all node embeddings.} We train our GNNs for 50 epochs on the aforementioned fusion group and buffer size range using an 80\%/10\%/10\% train/validation/test set split. GNN training takes a few minutes and both the LBDF and BRR GNNs achieve over 95\% classification accuracy and F1 performance on their respective test sets. Furthermore, we use the same training sets to pre-train the additional parameters required by the PG and RG GETs for 25 and 10 epochs, respectively, which takes a few hours.

\subsection{Scope of Layer Fusion Optimization}

We consider three search algorithms: Random Search (RS)~\cite{li2020random}, Local Search (LS)~\cite{stutzle1999local} and Non-dominated Sorting Genetic Algorithm II (NSGA-II)~\cite{yusoff2011overview}. RS is simple and unguided. As the name suggests, it randomly generates a new partition plan at every step, then determines its validity and DRAM access cost while tracking the best valid plan with the lowest access cost. LS is a mutation-driven Evolutionary Algorithm which maintains a fixed population of the top-$K$ best partition plans it has observed. At each iteration it makes a 1-edit random mutation to each partition plan and observes the change in validity and DRAM cost. These mutations consist of altering the existing fusion groups; e.g., shifting operations from one group to another, merging or splitting groups, etc. Finally, NSGA-II is a advanced evolutionary algorithm that incorporates crossover between existing plans in the top-$K$ population alongside random mutation. 

Each search algorithm is primarily parameterized by a fixed search budget, which determines the number of partition plans it can generate. We set different budget values depending on experimental setup (e.g., CNN architecture and buffer size) but keep it consistent between RS, LS and NSGA-II. Moreover, we set $K=10$ for LS and NSGA-II. We use OpenBox~\cite{li2021Openbox} to implement NSGA-II.

We incorporate our method into these search algorithms by using it to enhance their ability to generate partition plans. Whenever these schemes generate an invalid fusion group, they will simply attempt to randomly split it into a set of smaller, valid groups and do not consider DRAM access costs when doing so. We augment this process by first attempting to intelligently split the invalid fusion group in a cost-conscious manner using the recursive approach from Section~\ref{sec:method} and a given GET, e.g., GNNE, PG or RG. 

Additionally, we improve the efficiency of our scheme by using memoization. We cache the results of our recursive splitting scheme, e.g., using the original invalid fusion group as a key, and the optimized result as a value, to avoid redundant recomputations. Furthermore, we separately cache results from $F_\beta$ and $F_D$ for each individual fusion group encountered during search. 

Finally, we report the DRAM access cost amongst fuseable operations. That is, since LBDF requires that the weights of all operations be stored in on-chip memory for the entire execution of a fusion group, we identify unfusable operations, e.g., large convolutions in ResNet-101, and remove them from consideration prior to optimization. Since these operations are not part of the search space and must execute under another form of LF, we do not report their DRAM access costs and instead report metrics amongst fuseable operations.

\subsection{Improving Search on Large Networks}
\label{sec:results_large}

\begin{table*}[t]
    \centering
    \caption{DRAM access costs found for EfficientNet-B3, ResNet-152 and DeepLabV3+MobileNetV3 using Line Buffer Depth-First (LBDF) execution with an on-chip buffer size of 256KB. We implement GNNE, PG and RG with Local Search, NSGA-II and Random Search. We also report the Maximum Buffer Usage (MBU) of the corresponding layer fusion plan. Best results in bold. Lower DRAM access and higher MBU are preferred.}
    \label{tab:eb0_r152}
    \scalebox{0.95}{
    \begin{tabular}{lcccccc} \toprule
    & \multicolumn{2}{c}{\textbf{EfficientNet-B3}} & \multicolumn{2}{c}{\textbf{ResNet-152}} & \multicolumn{2}{c}{\textbf{DeepLabV3+MobileNetV3}} \\ \midrule
    \textbf{Search Method} & \textbf{DRAM Access} & \textbf{Max. Buffer Usage} & \textbf{DRAM Access} & \textbf{Max. Buffer Usage} & \textbf{DRAM Access} & \textbf{Max. Buffer Usage} \\ \midrule
    Local Search         & 90.500MB & 248.674KB & 80.733MB & 253.750KB & 116.644MB & 252.461KB \\
    Local Search + GNNE  & 78.007MB & 254.240KB & \textbf{74.461}MB & 199.875KB & 114.440MB & 252.461KB \\
    Local Search + PG    & \textbf{73.569}MB & 250.361KB & 76.421MB & 213.375KB & 114.180MB & 252.461KB \\
    Local Search + RG    & 78.433MB & 254.240KB & 75.245MB & 213.375KB & \textbf{113.942}MB & 252.461KB \\ \midrule
    NSGA-II              & 77.334MB & 249.236KB & 77.205MB & 253.375KB & 114.351MB & 252.461KB \\
    NSGA-II + GNNE       & \textbf{61.265}MB & 254.248KB & 68.581MB & 199.875KB & \textbf{113.602}MB & 252.461KB \\
    NSGA-II + PG         & 61.792MB & 254.217KB & 67.013MB & 213.375KB & 113.840MB & 252.461KB \\
    NSGA-II + RG         & 61.535MB & 254.218KB & \textbf{66.621}MB & 253.375KB & \textbf{113.602}MB & 252.461KB \\ \midrule
    Random Search        & 172.802MB & 251.674KB & 113.661MB & 207.750KB & 148.153MB & 235.156KB \\
    Random Search + GNNE & 169.928MB & 249.236KB & 96.021MB & 188.250KB & 143.925MB & 252.461KB \\
    Random Search + PG   & 164.086MB & 249.236KB & 96.413MB & 208.750KB & \textbf{142.548}MB & 252.461KB \\
    Random Search + RG   & \textbf{162.774}MB & 249.236KB & \textbf{94.453}MB & 213.375KB & 147.022MB & 236.141KB \\ \bottomrule
    \end{tabular}
    }
\end{table*}

To start, we consider two large Image Classification networks, EfficientNet-B3 and ResNet-152, as well as a Semantic Segmentation network consisting of a MobileNetV3 (MBv3)~\cite{howard2019searching} feature extractor and DeepLabV3 (DLv3)~\cite{deeplabv3plus2018} prediction head. We aim to find the partition plan which minimizes DRAM access, reported in megabytes (MB), in the LBDF context for a 256KB on-chip buffer. We also report the Maximum Buffer Usage (MBU) in kilobytes (KB). MBU corresponds to the size of the largest fusion group within the 
layer fusion plan\footnote{Although higher values of MBU are better, it cannot exceed the buffer size. While maximizing MBU is not an objective, it provides an additional facet of context to compare search performance.}.

We set a budget of 5k layer fusion partition plans per search. Table~\ref{tab:eb0_r152} reports our findings. First, we note that our GET-driven scheme always finds a better schedule than the baseline - DRAM access cost never increases, and in some cases we observe sizeable access cost savings. Specifically, on EfficientNet-B3, we can reduce DRAM access by over 10MB by pairing any search algorithm with a given GET. In fact, when using LS or NSGA-II we reduce DRAM access by over 15MB or 20\% compared to the baseline. Furthermore, we also observe DRAM reductions of over 5MB ResNet-152, and almost reduce access by 20MB using RS. DRAM reduction is smallest on DLv3+MBv3. However, this is expected since DLv3+MBv3 is a Semantic Segmentation network which processes higher-resolution images\footnote{EB3 uses $300^2$, DLv3+MBv3 uses $513^2$. Other ONNX models use $224^2$.}, which in turn increases the buffer memory requirement for each operation and further restricts the search space by reducing the number of LBDF-fuseable operations using a 256KB buffer. Nevertheless, using any GET yields a superior partition plan, which verifies the utility of our GET-based method and greedy tree-based recursive splitting algorithm.

When comparing across different search algorithms, our results corroborate intuitive expectations. That is, NSGA-II is the most advanced algorithm and obtains the best performance on each network, followed by LS. RS is the simplest and obtains the worst results, e.g., it cannot find a partition plan on ResNet-152 with DRAM access below 100MB without the assistance of our GET-driven method for splitting invalid fusion groups. Furthermore, we note that although NSGA-II and LS are guided search algorithms that have some understanding of what constitutes a low-cost partition plan or valid fusion group, they can still be improved by leveraging our GET-driven method to find lower-cost partition plans. Therefore, overall, these results demonstrate the generalizablity of our scheme when paired with different search algorithms. 

\begin{figure}[t]
    \centering
    \subfloat[GNNE]{\includegraphics[width=0.75in]{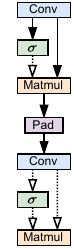}}
    \qquad
    \subfloat[PG]{\includegraphics[width=0.75in]{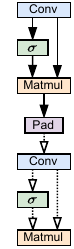}}
    \qquad
    \subfloat[RG]{\includegraphics[width=0.75in]{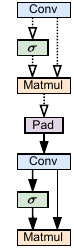}}
    \caption{Explanations of an invalid fusion group from EfficientNet according to GNNE, PG and RG. Solid lines indicate an edge was selected via by a given GET.}
    \label{fig:eb0_example}
\end{figure}

Next, we compare MBU results. We note how our recursive splitting can find fusion groups with larger buffer usage in many scenarios, such as EfficientNet-B3 for LS and NSGA-II. We also note how the MBU for DLv3+MBv3 is almost always the same value of 252.461KB. This likely corresponds to a large fusion group which most search algorithms (except the RS baseline) can easily find.

Finally, Figure~\ref{fig:eb0_example} provides example illustrations of the explanatory subgraphs generated by GNNE, PG, and RG on an invalid EfficientNet-B0 fusion group. While each GET selects different edges, we note some intuitive commonalities between them: they all select a major skip-connection (Conv to Matmul) which is not friendly to LBDF execution~\cite{shi2021hardware} as well as a padding operation.

\subsection{Comparisons Across Fusion Methods}

We examine whether GETs can improve search results across different buffer fusion methods using a 2k partition plan budget per search and a 128KB buffer size. Table~\ref{tab:main_local} tabulates our findings for several DNN networks using LS. We note how, in the BRR context, the use of our GET-driven method always yields partition plans with lower DRAM access costs. The same holds when considering LBDF LF, except on VGG16 and ResNet-18 where LS finds always finds a minima value as the sheer simplicity of both CNNs which enables LS to find the minima cost through brute-force.

Our findings in Table~\ref{tab:main_random} which considers RS, corroborate this explanation as each GET outperforms the baseline on the same partition plan budget. Overall though, we do observe smaller DRAM savings compared to those in Section~\ref{sec:results_large}. This is an expected outcome as smaller neural networks correspond to smaller LF optimization search spaces which are easier to optimize. By contrast, we observe more appreciable gains on the most complex DNN, MobileNetV2, where we can reduce DRAM access by 10\% or more compared to the baseline. Thus, the findings in Tables~\ref{tab:main_local} and \ref{tab:main_random} demonstrate how our method is generalizable to different forms of LF as well as search algorithms.

\begin{table*}[t]
    \centering
    \caption{DRAM Access in MB for MobileNetV2, VGG16, ResNet-50/18 and SqueezeNet under the BRR and LBDF settings with a 128KB on-chip buffer. Lower access is better. For this experiment, we consider a simple Local Search Algorithm and augment it with several GEs such as GNNExplainer, PGExplainer and RG-Explainer. Best results in bold.}
    \label{tab:main_local}
    \scalebox{0.95}{
    \begin{tabular}{l|ccccc|ccccc} \toprule
    & \multicolumn{5}{c|}{\textbf{BRR}} & \multicolumn{5}{c}{\textbf{LBDF}}  \\ \midrule
    Search & MobileNetV2 & VGG16 & ResNet-50 & ResNet-18 & SqueezeNet &  MobileNetV2 & VGG16 & ResNet-50 & ResNet-18 & SqueezeNet \\ \midrule
    LS        & 15.101 & 27.606 & 83.685 & 27.127 & 8.202 & 9.803 & \textbf{5.973} & 25.764 & \textbf{5.974} & 3.839 \\
    LS + GNNE & \textbf{13.401} & 27.415 & 83.344 & 27.127 & \textbf{6.232} & \textbf{8.823} & \textbf{5.973} & 24.588 & \textbf{5.974} & \textbf{3.585} \\
    LS + PG   & 14.609 & \textbf{26.975} & 83.493 & \textbf{26.975} & 6.790 & 9.068 & \textbf{5.973} & 24.588 & \textbf{5.974} & \textbf{3.585} \\
    LS + RG   & 13.790 & 27.174 & \textbf{83.317} & 27.127 & 7.714 & \textbf{8.823} & \textbf{5.973} & \textbf{24.196} & \textbf{5.974} & \textbf{3.585} \\ \bottomrule
    \end{tabular}
    }
\end{table*}

\begin{table*}[t]
    \centering
    \caption{DRAM Access in MB for MobileNetV2, VGG16, ResNet-50/18 and SqueezeNet under the BRR and LBDF settings with a 128KB on-chip buffer using Random Search augmented with GETs. Lower access is better.} 
    \label{tab:main_random}
    \scalebox{0.95}{
    \begin{tabular}{l|ccccc|ccccc} \toprule
    & \multicolumn{5}{c|}{\textbf{BRR}} & \multicolumn{5}{c}{\textbf{LBDF}}  \\ \midrule
    Search & MobileNetV2 & VGG16 & ResNet-50 & ResNet-18 & SqueezeNet &  MobileNetV2 & VGG16 & ResNet-50 & ResNet-18 & SqueezeNet \\ \midrule
    RS        & 16.195 & 29.274 & 91.866 & 28.525 & 10.249 & 13.209 & 6.856 & 39.092 & 6.856 & 10.080 \\
    RS + GNNE & \textbf{12.593} & 28.593 & \textbf{91.669} & 27.919 & 8.258 & \textbf{11.432} & \textbf{6.268} & 35.564 & \textbf{6.464} & 7.948 \\
    RS + PG   & 13.307 & \textbf{28.124} & 91.763 & \textbf{27.599} & 7.817 & 13.049 & 6.562 & 32.820 & 6.562 & 7.751 \\
    RS + RG   & 12.838 & 28.847 & 92.376 & 28.081 & \textbf{7.601} & 12.890 & 6.464 & \textbf{32.036} & \textbf{6.464} & \textbf{7.333} \\ \bottomrule
    \end{tabular}
    }
\end{table*}

\begin{figure*}
    \centering
    \subfloat[SqueezeNet on BRR using LS]{\includegraphics[width=2.15in]{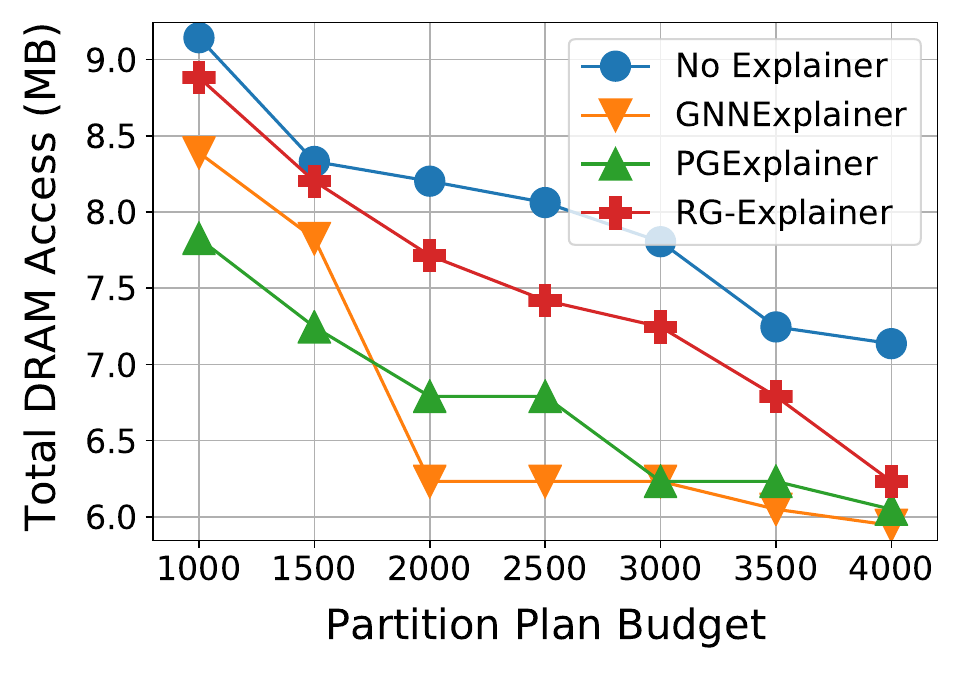}}
    \qquad
    \subfloat[MBv2 on LBDF using LS]{\includegraphics[width=2.15in]{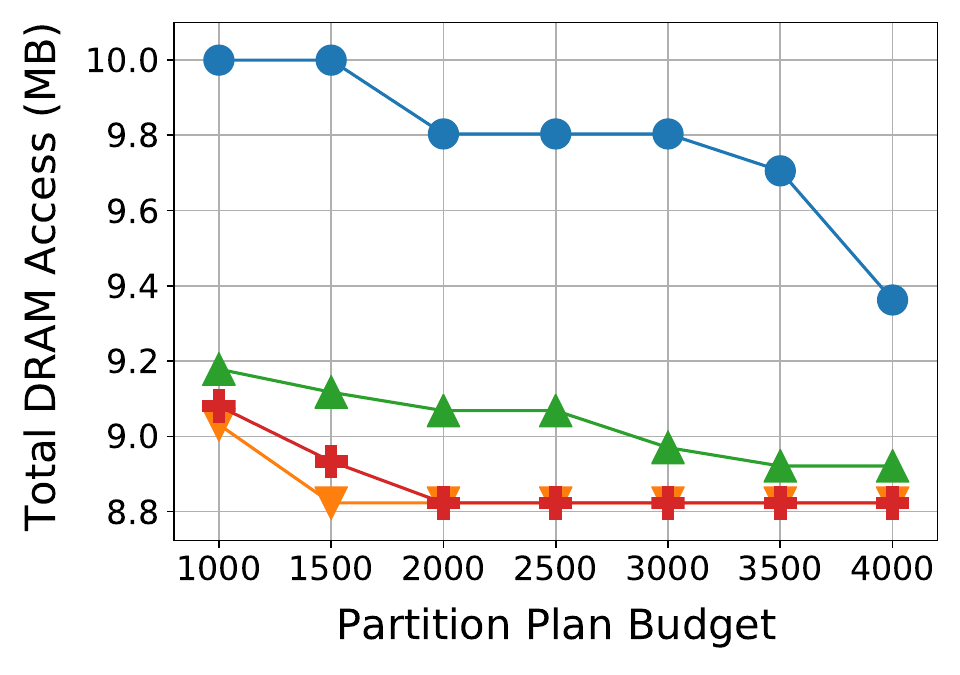}}
    \qquad
    \subfloat[VGG on LBDF using RS]{\includegraphics[width=2.15in]{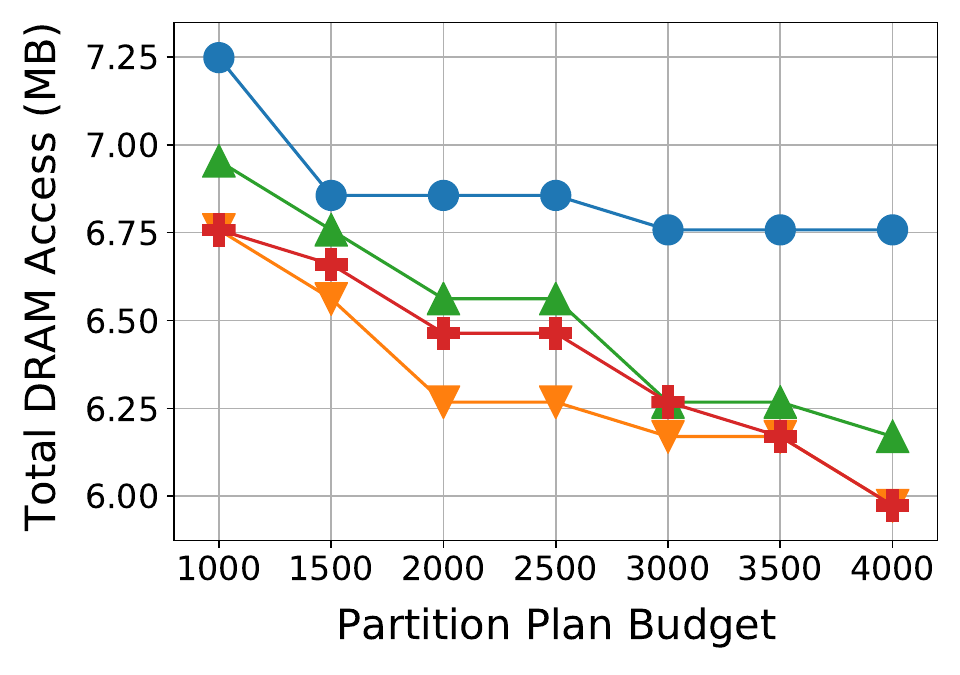}}
    \caption{Partition plan budget vs. best DRAM access cost. We compare DRAM performance across gradual increases in the plan budget. Best viewed in color.} 
    \label{fig:costs}
\end{figure*}

\subsection{Additional Figures and Discussion}
\label{sec:ablate}

To highlight the effectiveness of our GET scheme in terms of DRAM cost and evaluation budget, we run additional experiments where we vary the search partition plan budget. Figure~\ref{fig:costs} plots the results. We observe that the search methods enhanced by GETs consistently outperform the baselines at every budget value in each case. Moreover, Fig.~\ref{fig:costs}(b) shows that there are some cases where a GET with a 1k partition plan evaluation budget can outperform a baseline with a budget of 4k evaluations.

Next, we compare the effectiveness of each GET in terms of the ability to fix invalid fusion groups. Recall that, given an invalid group, a GET will return a set of edges that can be used to split the fusion group, hopefully into two smaller, valid groups. Specifically, given an invalid fusion group, we define the \textit{rectify rate} as the number of times an initial invalid fusion group was successfully broken up into a set of smaller subgraphs according to the explanatory edges the GE provides. We express this metric as a percentage and report it alongside wall-clock time, noting that the classification GNN, recursive splitting algorithm (Sec.~\ref{sec:method}), target network, and hardware are held constant. 

We report our results in Table~\ref{tab:explainer_compare}. We observe that while all three GETs can correct invalid fusion groups over 50\% of the time, GNNE and RG achieve noticeably higher rectify rates than PG. Next, we observe that PG and GNNE are the most and least efficient GETs in terms of time cost, respectively. This makes sense as GNNE~\cite{ying2019gnnexplainer} does not add any trainable parameters on top of the initial GNN and, therefore, must execute from scratch for each graph one wishes to explain. By contrast, both PG~\cite{luo2020parameterized} and RG~\cite{shan2021reinforcement} extend the concepts of GNNE by pre-training additional parameters on top of the initial GNN classifier in order to improve downstream runtime. Finally, RG is slightly slower than PG as it is based in Reinforcement Learning and builds a subgraph explanation over several sequential steps. 

Overall, our experimental results in this paper demonstrate the utility of GETs in LF. Comparing the different GETs against each other, both GNNE and RG tend to be superior to PG most of the time as they usually find lower DRAM costs (e.g., Tabs.~\ref{tab:eb0_r152}-\ref{tab:main_random} and Fig.~\ref{fig:costs}(b-c)) and achieve a higher rectify rate. Moreover, a trade-off exists between GNNE and RG. While RG achieves a higher rectify rate and lower wall-clock time for search, an additional pre-training step is required to learn the parameterized weights in addition to a pre-trained GNN, whereas the GNNE explanation process only requires a GNN but executes from scratch on each new graph instance. We give RG the advantage as the DRAM savings granted by having better partition plans likely outweigh the one-time cost of RG pretraining, especially since no additional data is required beyond what is used to train the initial GNN. 

%% file: src/conclusion.tex
\begin{table}[t]
    \centering
    \caption{Comparing various GETs in terms of how frequently they can correct an invalid fusion group (Rectify Rate) and search cost in seconds for MBv2 with LBDF.}
    \label{tab:explainer_compare}
    \begin{tabular}{lcc} \toprule
    \textbf{Search on MBv2 LBDF} & \textbf{Rectify Rate} & \textbf{Search Time} \\ \midrule
    Local Search + GNNE & 83.1\% & 168s \\
    Local Search + PG   & 59.1\% & 72s \\
    Local Search + RG   & 94.0\% & 99s \\ \midrule
    Random Search + GNNE & 72.1\% & 106s \\
    Random Search + PG   & 50.7\% & 32s \\
    Random Search + RG   & 91.4\% & 42s \\ \midrule
    \end{tabular}
\end{table}

\section{Conclusion}
\label{sec:conclusion}

We approach the problem of Layer Fusion (LF) optimization by applying Graph Explanation Techniques (GET) to improve search. GETs take an invalid fusion group and GNN as input and provides an explanation for why the fusion group cannot fit on a given on-chip buffer. We pair these GETs with a recursive partitioning method to split invalid fusion groups in a cost-conscious manner to minimize DRAM access. We consider modern and classical DNN designs such as EfficientNets, MobileNets, ResNets and SqueezeNets for Image Classification and Semantic Segmentation in the LBDF and BRR LF scenarios. We pair our method with off-the-shelf search algorithms such as Local Search, NSGA-II and Random Search which show that a broad range of search algorithms can utilize our method to augment the optimization process. Experimental results show that our proposed scheme is effective at splitting invalid fusion groups while minimizing DRAM cost. For example, we can substantially reduce DRAM access on large classification architectures like EfficientNet-B3 and ResNet-152, where we reduce access cost by over 15MB and 20MB, respectively, compared to the baseline. Moreover, we demonstrate the efficiency of our scheme by showing how it can find better layer fusion partition plans with lower search budgets.